\documentclass{article}

\usepackage{arxiv}

\usepackage[utf8]{inputenc} 
\usepackage[T1]{fontenc}    
\usepackage{hyperref}       
\usepackage{url}            
\usepackage{booktabs}       
\usepackage{amsfonts}       
\usepackage{nicefrac}       
\usepackage{microtype}      
\usepackage{lipsum}
\usepackage{graphicx, subcaption}

\usepackage[numbers]{natbib}

\title{Finding manoeuvre motifs in vehicle telematics}

\author{
  Maria Inês Silva \\
  Nova Information Management School (NOVA IMS)\\
  Campus de Campolide, Universidade Nova de Lisboa\\
  1070-312 Lisboa, Portugal \\
  \texttt{d20170088@novaims.unl.pt} \\
   \And
 Roberto Henriques \\
  Nova Information Management School (NOVA IMS)\\
  Campus de Campolide, Universidade Nova de Lisboa\\
  1070-312 Lisboa, Portugal \\
  \texttt{roberto@novaims.unl.pt} \\
}

\begin{document}
\maketitle

\begin{abstract}
Driving behaviour has a great impact on road safety. A popular way of analysing driving behaviour is to move the focus to the manoeuvres as they give useful information about the driver who is performing them. In this paper, we investigate a new way of identifying manoeuvres from vehicle telematics data, through motif detection in time-series. We implement a modified version of the \textit{Extended Motif Discovery} (EMD) algorithm, a classical variable-length motif detection algorithm for time-series and we applied it to the UAH-DriveSet, a publicly available naturalistic driving dataset. After a systematic exploration of the extracted motifs, we were able to conclude that the EMD algorithm was not only capable of extracting simple manoeuvres such as accelerations, brakes and curves, but also more complex manoeuvres, such as lane changes and overtaking manoeuvres, which validates motif discovery as a worthwhile line for future research.
\end{abstract}


\section{Introduction}

Driving behaviour has a great impact on road safety. According to \citet{dingus_driver_2016}, accident causation has experienced a shift, with driver-related factors being a fundamental factor for road safety. On a similar study, \citet{dozza_what_2013} found that some behaviours associated with distraction, such as attendance to secondary tasks and eyes-on-road, reduce significantly drivers' response time in near-accident settings and, therefore, distraction and inattention are a major factor in traffic accidents. 

On the other hand, because data acquisition systems such as smartphones or on-board diagnostics (OBD) devices are easier to access and data is cheaper to store and process, researchers have the opportunity of using vehicle telematics to better understand driving behaviours and the factors contributing to car accidents \citep{carsten_vehicle-based_2013}. 

Driving behaviour analysis also has impact for practitioners. In the insurance market, driving behaviour has been used through the implementation of usage-based motor insurance schemes (UBI) \citep{tselentis_innovative_2017}. These schemes rely on collecting continuous streams of cars’ sensor data on single customers and using that data to assess their driving behaviour and provide discounts on the premiums. Other applications include fuel consumption optimisation, fleet management and evaluation of self-driving cars' performance.

When tackling the driving behaviour problem, it is common to focus on manoeuvres. The exact manoeuvres being performed during a trip and the way they are being performed is very informative of the driving behaviour of the driver. In this context, the question of how to use telematics to correctly identify and compare manoeuvres needs to be addressed. Previous research has dealt this issue through two main strategies, namely, 1) using fixed thresholds in inertial measurements (e.g. acceleration or rotational energy) to define the start and end of specific manoeuvres or 2) using features extracted from rolling windows of sensor data (e.g. velocity and acceleration) in a supervised learning model to detect manoeuvres. While the first method cannot adapt to small fluctuations in the signal and requires fine-tuning, the second requires a dataset with labels indicating where the manoeuvres appear and cannot identify manoeuvres with different lengths in time.

Having in mind the need for an adaptable method that can detect manoeuvres without the need of labels, we sought to investigate a type of methods created in the time-series data mining community that can bring insights into the manoeuvre detection use-case, namely, the algorithms for detecting time-series motifs. Particularly, after a review of the available algorithms for motif detection, we made our own implementation of a specific algorithm, applied it to a publicly available naturalistic driving dataset and explored the relationship between the extracted motifs and manoeuvres.

The rest of the paper is organised as follows - section 2 includes a review on the methods for identifying driving behaviour with telematic data and explores the field of time-series motif detection algorithms, section 3 presents the specific algorithm used for this paper and explains the modifications added to better fit this use-case, section 4 summarises the main results and the fifth and final section concludes the paper and opens the path for future research.

\section{State of the art}

\subsection{Driving behaviour in telematic data}

Identifying driving behaviour or detecting manoeuvres high-frequency trip recordings is not a straightforward task and many authors proposed some ways of tackling this problem. Nevertheless, most papers can be aggregated into two main strategies - fixed thresholds \citep{johnson_driving_2011, paefgen_driving_2012, kantor_design_2014, eren_estimating_2012} on inertial variables and rolling windows \citep{saleh_driving_2017, weidner_classification_2016, murphey_drivers_2009, xie_driving_2018, singh_smartphone_2017, woo_manoeuvre_2016, camlica_feature_2016, wu_novel_2016, junior_driver_2017}.

In the fixed thresholds strategy, authors set specific thresholds on acceleration or other inertial measurements in order to define the beginning and end of manoeuvres. These manoeuvres are then used as input to the driving performance algorithm. As an example, \citet{paefgen_driving_2012} defined four types of manoeuvres by specifying thresholds on lateral acceleration (for the left turn and the right turn) and on longitudinal acceleration (for forward acceleration and for braking). In another paper, \citet{johnson_driving_2011} applied a simple moving average on the rotational energy derived from smartphone's sensors and defined a specific interval to set the beginning and end of a manoeuvre, independently of its type. These types of approaches have the advantage of being easy to implement, lightweight and interpretable. However, since the choice of the thresholds is essential to the detection of manoeuvres, these methods require fine-tuning for each specific dataset and are inflexible to changes in the data.

The second type of strategies involves the use of rolling windows, in which the trip is divided into fixed-sized time windows that can have some level of overlap. Then, each window is used as input to a supervised model that classifies it as a specific manoeuvre. In this strategy, the window size and the percentage of overlap are important parameters that need to be set beforehand. \citet{saleh_driving_2017} built windows from inertial sensors and classified them as aggressive, normal or drowsy using a stacked-LSTM model. Each window had 64 time steps and had a 50\% overlap. On a different direction, \citet{xie_driving_2018} compared three feature extraction methods for identifying various manoeuvres such as braking, turning and acceleration. They also used rolling windows for splitting the trips and they tested different window sizes with no overlap. The authors state that the window size is very relevant to the classification performance and that the optimal size can vary depending on the type of manoeuvre, which leads to the conclusion that an "adaptive window sizing method" would be better suited. Thus, again, the inflexibility of the window size is a disadvantage in this method as well.

Another issue with the rolling windows method is the overlap parameter. If no overlap is used, then the way a trip is split may not be optimal since a window can also split a single manoeuvre and information may be lost. On the other hand, using the maximal overlap (i.e., moving the window a single time step) comes with its own problems. In their paper, \citet{keogh_clustering_2005} argued that clustering time-series subsequences is meaningless when rolling windows are used to build the subsequences. In particular, they state that \textit{"clusters extracted from these time series are forced to obey a certain constraint that is pathologically unlikely to be satisfied by any dataset, and because of this, the clusters extracted by any clustering algorithm are essentially random"}.

It is important to note that \citet{saleh_driving_2017} used an 50\% overlap with great success, which seems to validate the use of rolling windows. However, the authors used a supervised approach and \citet{keogh_clustering_2005} only argues against using rolling windows in unsupervised methods. Nevertheless, building high-frequency labelled datasets for driving behaviour is quite laborious and requires a lot of resources, which is evident in the extremely low number of labelled datasets available to researchers. Thus, and since the possibility of using unsupervised or semi-supervised methods is quite desirable, for this specific problem, using rolling windows is not the optimal approach.

The paper from \citet{keogh_clustering_2005} not only proves the point of the meaninglessness of clustering rolling windows but also proposes a solution, \textit{motif-based clustering}. According to the authors, time-series motifs are over-represented subsequences in a time-series. The concept of motifs was first introduced in the genetics research as sequences of amino-acids in the DNA with biological significance \citep{torkamani_survey_2017} and, since then, the field of motif discovery for time-series has been receiving a lot of attention from the data mining community. \citet{keogh_clustering_2005} proposes to use a motif detection algorithm to find the set of motifs with the highest representation in the original time-series and then to apply the clustering algorithm directly to the motifs.

Therefore, in this paper, we propose to use a motif detection algorithm in inertial measurements to extract manoeuvres from a trip. In other words, our main hypothesis is that over-represented segments of inertial time-series are highly connected to manoeuvres and, by looking at the most relevant motifs for a trip, we will be able to get insights on the manoeuvres performed during that trip.

In the next subsection, we will explore the motif discovery task, taking into consideration the specific requirements of the manoeuvre detection use-case. The method needs to be flexible enough to detect manoeuvres with variable sizes, which implies that the motif detection algorithm needs to be able to extract variable-length motifs.

\subsection{Motif discovery overview}

Even though the concept of motifs relates to over-represented time-series segments, the exact definition of the most relevant motifs varies slightly among authors and areas of application. \citet{mueen_time_2014} states that there are two main definitions, namely a similarity-based and a support-based. In the first, the motifs can be ordered based on the similarity of the segments that belong to the motif, while the second orders motifs based on the number of repetitions of the motif throughout the time-series. Thus, the similarity-based definition results in highly similar motifs and the support-based definition results in highly frequent motifs.

Additionally, there is some constrains which the time-series segments belonging to a motif must meet \citep{tanaka_discovery_2005}. The behaviour constrain introduces the idea that segments in a motif should present the same general behaviour. Note that, depending on the application, distortions such as noise or time and amplitude shifts may be accepted \citep{torkamani_survey_2017}. The distance constrain goes a bit beyond by stating that all motif's segments need to have a distance smaller than a predefined radius $R$ to the centre of the motif (i.e., the segment that represents that motif). This radius is a very important parameter in any motif detection algorithm and it needs to set beforehand. Finally, the non-overlapping constrain sets that motif segments cannot overlap in time, which avoids what \citet{lin_finding_2002} refers to as trivial matchings.

There are also some motif detection use-cases in which one is not interested in finding the single most important motif, but instead the $k$ most important motifs or, as referred in the literature, the \textit{$k$-motifs} \citep{lin_finding_2002}. Put simply, based on the desired motif definition, one can order the entire set of motifs found with a certain algorithm and then extract the first $k$ ordered motifs. However, in order to avoid extracting motifs with overlapping members, these motifs have to meet a distance constrain, namely, all the pairs of motif's representative (or the centres) need to have a distance higher than $2R$, where $R$ is the radius discussed in the previous paragraph.

Independently of the exact definition, motif discovery is a computational intensive task as it involves comparing all possible pairs of time-series subsequences. The most prevailing strategy to speed this search is to use a low dimensional representation of the original time-series in which the distance of subsequences in the original representation is approximately maintained \citep{mueen_time_2014}. This way, one is able to extract motif candidates in the low-dimensional representation (which is more efficient than use the original representation) and simply filter the candidates based on the real distance in the original representation. The most used representation for motif detection is the \textit{Symbolic Aggregate approXimation} (SAX), which transforms sliding windows of the original time-series into discrete sequences of characters \citep{lin_experiencing_2007}. Since each sliding window is encoded with a predefined number of characters, the continuous time-series is broken into a list of fixed-sized sequences of characters (or words). Another common approach is to used spectral representations such as the \textit{Discrete Fourier Transformation} (DFT), the classical representation of a time-series in the frequency domain \citep{agrawal_efficient_1993}.

Another important component of motif discovery is the distance used to compare time-series segments. Most papers use either the Euclidean distance \citep{das_rule_1998} or the \textit{Dynamic Time Warping} distance (DTW) \cite{berndt_using_1994}. While the first is very efficient and thus allows for a fast comparison of segments, the second has the advantage of allowing to compare segments with different lengths and with time distortions by computing the optimal alignment in time of the two time-series that are being compared.

Motif discovery in time-series started with the fixed-length motif problem, where the goal was to find motifs with a predefined length. However, as stated in the introduction, we are interested in finding motifs with no predefined length and even allowing segments from the same motif to have different lengths. Therefore, because the fixed-length motif discovery is too restrictive for this use-case, we will need to use variable-length motifs.

According to \citet{torkamani_survey_2017}, there is a considerable set of papers that address the variable-length problem by applying a fixed-length algorithm to a range of window sizes and then choosing the most representative motifs based on their motif definition and motif ranking schemes \citep{nunthanid_parameter-free_2012, gao_exploring_2018}. However, this strategy does not allow to have segments with different lengths in the same motif. On the other hand, \citet{tanaka_discovery_2005} introduced a strategy that allows a more flexible comparison of motifs with different lengths. In particular, they proposed to use the SAX representation for building a discrete list of words and to aggregate repeating words in the same word.

\section{Implementation of the motif detection algorithm}

The motif discovery algorithm chosen for this paper is the \textit{Extended Motif Discovery} (EMD) algorithm by \citet{tanaka_discovery_2005}. The main advantage of this algorithm, and the reason why we chose it, is the possibility of having subsequences with different lengths in the same motif. The EMD algorithm has three main components:

\begin{itemize}
    \item Discretization of the 1-dimensional time-series via an adaptation of SAX representation \citep{lin_experiencing_2007}. This adaptation is what allows the method to find segments with different lengths in the same motif. Put simply, the algorithm starts by applying the SAX representation to sliding windows of the same length, which produces a sequence of SAX words, each representing a specific window in the original time-series. Then, the algorithm looks for consecutive sets of equal SAX words. If consecutive repeating words are found, the algorithm aggregates them into the same modified SAX word and joins the windows that were represented by those repeated words. For instance, given the SAX sequence $C_{SAX} = [abc, abc, cde, fde, fde, fde]$ representing 6 sliding windows of size $n$, the adaptation returns the modified sequence $\tilde{C}_{SAX} =[abc, cde, fde]$ that represents 3 windows of size $n+1$, $n$ and $n+2$, respectively. 
    \item Extraction of all the variable-length motif candidates with an iterative pattern matching routine. This process iterates over all the sets of consecutive modified SAX words (or patterns) and looks for repeating patterns that meet the distance constrain discussed in previous section.
    \item Computation of the description length of each motif candidate, which is based on the Minimum Description Length (MDL) principle \citep{rissanen_stochastic_1998}. The description length (or MDL cost) can be then used to select the most relevant motifs from the extracted candidates.
\end{itemize}

Since, to the best of our knowledge, there was not a publicly available implementation of the EMD algorithm, we did our own implementation of the algorithm (\textit{footnote with link to github}). In addition, implementing the algorithm gave us the the flexibility to add two simple features that were not discussed in the original paper and a major change that we will argue benefits the use-case of finding manoeuvres from inertial measurements.

The first addition is related to the avoidance of the trivial matchings discussed in the previous section. In their paper, \citet{tanaka_discovery_2005} did not give any details on how to prune overlapping subsequences that belong to the same motif. Therefore, we implemented it with a simple heuristic. Given a motif, its centre subsequence and its members subsequences, when two members overlap, we exclude the member with the largest distance to the motif's centre and keep the member with the lowest distance.

The second addition involves the extraction of $k$-motifs. Since we are interested in finding different manoeuvres, we had to implement a routine that pruned the entire set of motifs found by the EMD algorithm in order to avoid overlaps between different motifs. \citet{tanaka_discovery_2005} had already defined a method for ordering motifs using the MDL cost. However, they did not discuss the issue of overlapping motifs and thus we used the heuristic proposed by \citet{lin_finding_2002}. Namely, we defined the $k$-motif as the motif with the lowest MDL cost and whose centre has a distance higher than $2R$ to the the centre of each the $j$-motifs, for $1 \leq j \leq k-1$. We named this step \textit{motif pruning}.

 The major change we introduced to the EMD algorithm concerns the SAX representation. In the original method, each sliding window is transformed into a sequence of $n$ numbers, which is defined by dividing the window into $n$ equal-size segments and by computing the mean of each segment. Then, each of the $n$ numbers is mapped to a unique character based a set of computed break-points. These break-points are defined separately for each window so that the characters' frequency in that window exhibit a Gaussian distribution.
 
 With this adaptive break-points method, windows that have the same behaviour but are shifted in amplitude are mapped to the same SAX word and consequently will be compared as candidates for the same motif. Although useful in some cases, because amplitude is extremely important for identifying different manoeuvres, this adaptive feature is not desirable in our use-case. As an example, if we used adaptive thresholds, acceleration increases from 0G to 0.2G would have the same SAX word as acceleration increases from 0G and 0.8G and thus, we wouldn't be able to have these two subsequences as distinct motifs. In other words, we would not observe smooth manoeuvres and aggressive manoeuvres as different motifs.

Finally, after some investigation of the results, we observed that the motif punning reduced significantly the number of motifs detected. In order to better investigate all the original motifs found by the EMD algorithm, we applied a clustering method to the centres of each motif and grouped them based on their DTW distance. For this purpose, we used the DBSCAN algorithm \citep{ester_density-based_1996}. This is a density-based algorithm that defines core points of high-density (points with many close neighbours) and expands clusters from these points. The algorithm has two main parameters, namely, the maximum distance to consider two points neighbours and the minimum number of neighbours a points must have to be considered a core points. Based on this two parameters, the algorithms finds both outliers and the clusters in the data. This is a main advantage as we do not need to give a predefined number of clusters and can discover outlier motifs, which can be also interesting to analyse.

\section{Results and discussion}

In order to investigate whether the EMD algorithm is able to extract meaningful motifs for detecting manoeuvres, we applied the algorithm to the UAH-DriveSet, a publicly available naturalistic driving dataset with trip recordings from six different drivers and two specific routes in Madrid, Spain. \citet{romera_need_2016} asked each driver to repeat two predefined routes simulating three different behaviours, namely, normal, aggressive and drowsy. During the trips, they collected  both raw and processed signals using an app designed by them and called DriveSafe \cite{bergasa_drivesafe:_2014, romera_real-time_2015}.

We run two experiments\footnote{https://github.com/misilva73/manueverMotifs}. In the first, we aimed to identify brakes and accelerations, while in the second we focused on curves and other lateral manoeuvres. For each experiments, we run the algorithm in four different trips - one representing a normal trip in a secondary road and three related to trips in a motorway and exhibiting the three behaviours present in the dataset (normal, aggressive and drowsy).

In the experiments, we used the accelerometer's measurements in the $y$ and $z$ axis with a frequency of 10Hz, which were already aligned with the three car axis and denoised with a Kalman filter. Since the $y$-axis represents the lateral acceleration of the car and the $z$-axis represents the longitudinal acceleration of the car, we used the first to detect curves and the second to detect accelerations and brakes. This dataset also included a record of the start of acceleration events captured by the DriveSafe app, such as turns, brakes and accelerations. These events were marked with a fixed-thresholds strategy and thus, we used these labels to help in the motif visualisation.

The EMD algorithm has four main parameters, namely, the window size used to build each individual SAX word, the Piecewise Aggregate Approximation (PAA) size, which corresponds to the number of characters in each SAX word, the SAX alphabet size, which is the maximum number of characters used in the SAX representation, and the radius $R$, discussed in the definition of a motif. We tested a range of values for each parameter and, based on an exploration of the resulting motifs, the final parameters used in all the experiments were \texttt{window\_size = 20}, \texttt{paa\_size = 2}, \texttt{alphabet\_size = 5} and \texttt{R = 0.1}.

In order to validate the results, we used the video recordings of the trips and the measurements of car's velocity. The videos were mostly useful in the lateral experiment, while the velocity was used to validate the longitudinal manoeuvres. Note that we did this in-depth analysis on the smaller set of pruned motifs as the full set of extracted motifs was too big for such manual analysis. Therefore, we applied the clustering algorithm DBSCAN \citep{ester_density-based_1996} as an exploratory tool.

In order to explore all the motifs extracted

\subsection{Identifying brakes and accelerations in the longitudinal acceleration}

Table \ref{table:lon} summarises the main results obtained with the EMD method and the DBSCAN clustering. For each trip, it displays 1) the number of motifs extracted by the EMD algorithm, 2) the number of motifs remaining after the motif pruning, 3) the types of manoeuvres detected after punning, 4) the number of clusters obtained with the DBSCAN algorithm, excluding the outliers cluster, and 5) the number of motifs in the outlier cluster.

\begin{table}[ht!]
    \centering
    \begin{tabular}{llccccc}
        \hline
        Trip &  & Motifs & \begin{tabular}[c]{@{}c@{}}Motifs after \\ pruning\end{tabular} & Manoeuvres$^{[*]}$ & \begin{tabular}[c]{@{}c@{}}DBSCAN\\ clusters\end{tabular} & \begin{tabular}[c]{@{}c@{}}DBSCAN\\ outliers\end{tabular}\\
        \hline
        Motorway & Normal & 1849 & 3 & B, B-A & 5 & 383\\
         & Aggressive & 1272 & 8 & B, B-A & 8 & 771\\
         & Drowsy & 2039 & 5 & B, B-A & 5 & 477\\
        Secondary & Normal & 1591 & 3 & A, B-A & 2 & 162\\
        \hline
    \end{tabular}
    \\
    \vspace{1 mm}
    \scriptsize{$[*]$ A = Acceleration ; B = Brake ; B-A = Brake followed by acceleration}
    \caption{Summary results of the analysis in the longitudinal acceleration}
    \label{table:lon}
\end{table}

\begin{figure}[ht!]
    \centering
    \begin{subfigure}[t]{0.7\linewidth}
        \includegraphics[width=\linewidth]{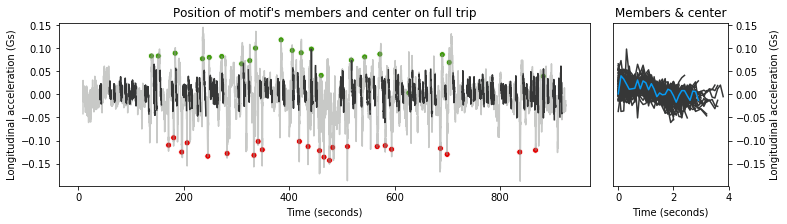}
        \caption{No manoeuvre}
    \end{subfigure}\vspace{4mm}
    \begin{subfigure}[t]{0.7\linewidth}
        \includegraphics[width=\linewidth]{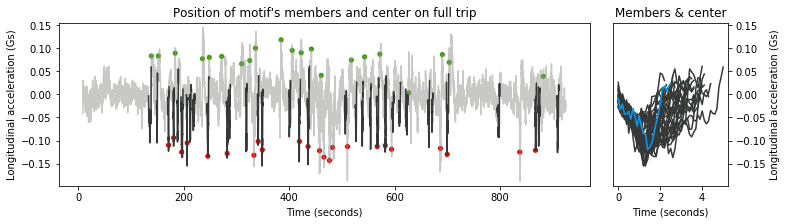}
        \caption{Brake manoeuvre}
    \end{subfigure}\vspace{4mm}
    \begin{subfigure}[t]{0.7\linewidth}
        \includegraphics[width=\linewidth]{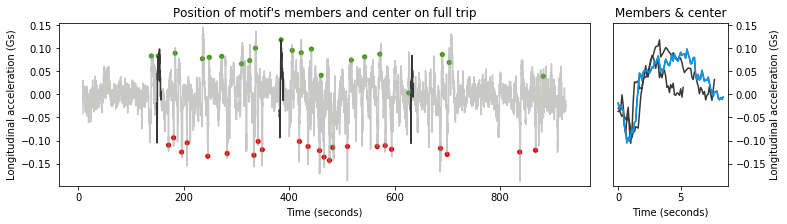}
        \caption{Brake-acceleration manoeuvre}
    \end{subfigure}
    \caption{Three example motifs found in the motorway aggressive trip. They correspond to the first, second and fifth most relevant motifs, after pruning. Red and green points represent the original labels (computed using a threshold method) present in the UAH-DriveSet, where red marks indicate the beginning of a brake and green the beginning of an acceleration.}
    \label{fig:lon}
\end{figure}

From these results, we were able to observe that the motif pruning step reduces the number of motifs by three orders of magnitude, which is a very significant reduction. This indicates that most of motifs detected by the EMD algorithm are in fact very close to each other and therefore represent either the same manoeuvre or very similar manoeuvres. Additionally, after the motif pruning step, the remaining motifs in all trips contained relevant manoeuvres, which indicates that motif detection algorithms can help in the task of discovering manoeuvres in a signal of longitudinal acceleration. Figure \ref{fig:lon} displays three motifs extracted with the EMD algorithm and they all have an easily identifiable manoeuvre.

In all the trips explored, the most relevant motif found by the EMD algorithm contained the absence of a manoeuvre. In other words, this motif corresponded to the action of driving at a constant speed without changes in longitudinal acceleration, which is consistent with routes where these trips where recorded. In motorways and secondary roads, driving is done in a constant speed at most times and thus we expect to have this behaviour as the most significant motif of these trips.

It is also interesting to note that in all the four trips, the algorithm was able to identify a motif with the manoeuvre of a brake followed by an acceleration, which could be indicative of a tailgating behaviour.

Another common factor to all the explored trips was the variability of the lengths of segments belonging to the same motif, which was one of the main reasons for choosing the EMD algorithm in the first place. The ability of extracting motifs whose members have slightly different lengths is a major advantage of this algorithm as it allows us to find manoeuvres that have the same type (e.g. a brake) but have lightly different lengths.

\begin{figure}[ht!]
    \centering
    \includegraphics[width=0.7\linewidth]{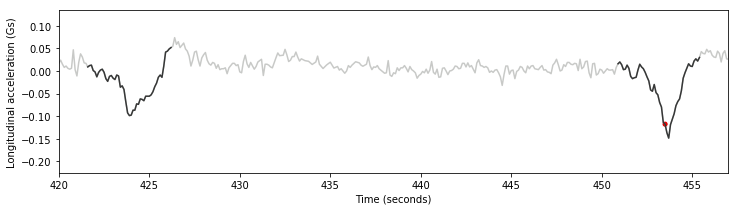}
    \caption{Zoom-in on a motif extracted from the drowsy trip in the motorway route. The second subsequence was marked as a brake in the UAH-DriveSet and the EMD algorithm extracted it as a member of a motif. The first, although not being marked, could be also recognised as a brake by belonging to the same motif as the first subsequence.}
    \label{fig:zoom}
\end{figure}

We also observed that this method was capable of finding subsequences that correspond to the same manoeuvre but, because they are slightly bellow the thresholds set in the DriveSafe app, they are not marked as a manoeuvre in the original dataset. As an example, figure \ref{fig:zoom} shows the plot of two subsequences that belong to the same motif. The first was marked as a brake but the second, because it was bellow the threshold set by the app, was not marked as a brake. This motif was extracted from the drowsy trip in the motorway route.

The results obtained with the DBSCAN clustering algorithm show that the majority of the motifs are very close and represent the same behaviour, the absent of a manoeuvre. However, all trips have a cluster of outliers, which are motifs that are different from the majority motifs and do not have a big enough neighbourhood to form a cluster. These are the most interesting motifs from the perspective of manoeuvre detection. Therefore, investigating outlier motifs can be a way of enriching the motifs found with the pruned version of the EMD algorithm.

\subsection{Identifying turns in the lateral acceleration}

\begin{figure}[ht!]
    \centering
    \begin{subfigure}[t]{0.7\linewidth}
        \includegraphics[width=\linewidth]{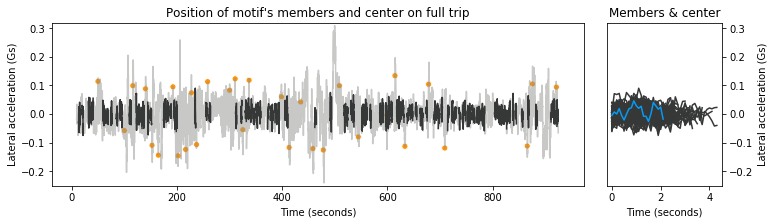}
        \caption{No manoeuvre}\vspace{4mm}
    \end{subfigure}
    \begin{subfigure}[t]{0.7\linewidth}
        \includegraphics[width=\linewidth]{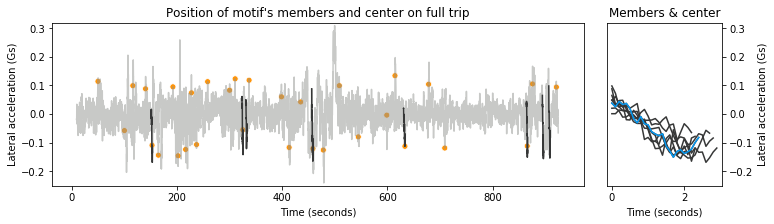}
        \caption{Lane change}\vspace{4mm}
    \end{subfigure}
    \begin{subfigure}[t]{0.7\linewidth}
        \includegraphics[width=\linewidth]{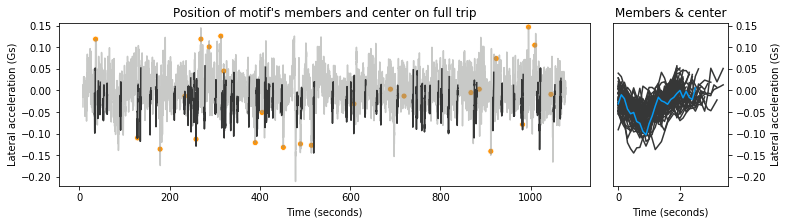}
        \caption{Drift}
    \end{subfigure}
    \caption{Two example motifs found in the motorway aggressive trip (1st and 4th motif) and one example found in the motorway drowsy trip (2nd motif). Orange points represent the original labels present in the UAH-DriveSet (and computed using a threshold method), where the marks indicate the beginning of a turn.}
    \label{fig:lat}
\end{figure}

Similarly to the previous section, table \ref{table:lat} summarises the main results obtained with for the lateral acceleration analysis and figure \ref{fig:lat} shows three motifs extracted with the EMD algorithm with pruning. In general, results were consistent with the ones obtained in the longitudinal acceleration experiments and thus conclusions are not much different.

\begin{table}[ht!]
    \centering
    \begin{tabular}{llccccc}
        \hline
        Trip &  & Motifs & \begin{tabular}[c]{@{}c@{}}Motifs after \\ pruning\end{tabular} & Manoeuvres$^{[*]}$ & \begin{tabular}[c]{@{}c@{}}DBSCAN \\ clusters\end{tabular} & \begin{tabular}[c]{@{}c@{}}DBSCAN \\ outliers\end{tabular} \\
        \hline
        Motorway & Normal & 1532 & 3 & LC, OT & 6 & 1328\\
         & Aggressive & 634 & 7 & D, LC & 1 & 614\\
         & Drowsy & 1464 & 8 & LC, C & 5 & 1292\\
        Secondary & Normal & 1113 & 1 & None & 2 & 929\\
        \hline
    \end{tabular}
    \\
    \vspace{1 mm}
    \scriptsize{$[*]$ LC = Lane change ; OT = Overtaking ; D = Drift, C = Curve}
        \caption{Summary results of the analysis in the lateral acceleration}
    \label{table:lat}
\end{table}

Motif pruning continues to reduce significantly the number of motifs and, after pruning, we could identify some relevant manoeuvres in the remaining motifs. In all trips, the most relevant motif was the one representing the absence of a manoeuvre which was expected given that both routes do not have many curves. In three trips, some of the motifs extracted after the pruning related to lane changes, which was easily validated in the videos recordings. Additionally, in one of the trips, we were able of observe a motif that included a mush more complex manoeuvre, namely a overtaking manoeuvre.

The video recordings also showed that some motifs included clear patterns in the lateral acceleration which, in most cases, were related to a lane change but, in a few cases, included no visible manoeuvre. We hypothesise that these patterns in the acceleration time-series can be caused by road inclinations or blasts of wind. This means that by looking at a single signal, motifs will be more prone to errors and thus, as future work, we should combine multiple sensors and find multi-dimensional motifs.

Compared to the longitudinal analysis, the DBSCAN detected more outliers, which suggests that this signal leads a higher number of motifs which are distant from the "no manoeuvre" behaviour and thus one would need to spend more time analysing these motifs.

\section{Conclusion and future work}

As many papers have shown, driving behaviour has a great impact on road safety. A popular way of analysing driving behaviour is to move the focus to the manoeuvres as they give useful information about the driver performing those manoeuvres. In this paper, we investigated a new way of identifying manoeuvres from vehicle telematics data, motif detection in time-series. Put simply, time-series motifs are over-represented subsequences in a time-series \citep{lin_finding_2002}. With the hypothesis that over-represented segments of inertial time-series are highly connected to manoeuvres, we sought to analyse the relationship between the most relevant motifs of a trip and the manoeuvres performed during that trip.

We implemented a slightly modified version of the \textit{Extended Motif Discovery} (EMD) algorithm \citep{tanaka_discovery_2005}, a classical motif detection algorithm for time-series which is capable of finding subsequences with different lengths in the same motif, and we applied it to the UAH-DriveSet \citep{romera_need_2016}, a publicly available naturalistic driving dataset with trip recordings from six different drivers and two specific routes in Madrid, Spain. Particularly, we ran two different experiments. In the first, we aimed to identify acceleration and brakes from the longitudinal acceleration time-series and, in the second, we aimed to identify turns from the lateral acceleration time-series.

After a systematic exploration of the extracted motifs, we were able to conclude that the EMD algorithm was capable of extracting simple manoeuvres such as accelerations, brakes and curves. We identified additional manoeuvres that could be associated with more complex behaviours. In the longitudinal experiments, we identified a motif with the manoeuvre of a brake followed by an acceleration, which could be indicative of a tailgating behaviour. In the lateral experiment, we found motifs with lane changes and overtaking manoeuvres.

Additionally, by providing a way of ordering the motifs by its relevance in the trips, the EMD algorithm gives some extra information on the trip itself. As an example, in the drowsy trip, the second most relevant manoeuvre in the lateral acceleration was the drift while, in the aggressive trip, the lane changes occupied the most interesting motifs. This is very indicative of the type of driving that was being made in both trips: the aggressive trip was dominated by lane changes, thus showing a higher level of impatience, and the drowsy trip was dominated by drifts, which can be associated with inattention and sleepiness.

Although validating motif discovery as a worthwhile line of research for detecting manoeuvres, there is still some work to be done. Firstly, the analysis performed in this paper was exploratory in nature as we did not have a more automated way of identifying which manoeuvre was being performed in each motif. Therefore, an important next step to make motif detection work in practice is to build a method for finding similar motifs and associating that motifs to a specific manoeuvre without the need to visual exploration.

Secondly, even though motif pruning showed some promise in ordering and selecting motifs, the DBSCAN analysis leads us to believe that by only looking at the pruned motifs and discarding all the others, we might be missing some interesting manoeuvres. Therefore, it would be also interesting to further investigate a better grouping of the motifs as a way of finding more instances of with manoeuvres.

Thirdly, we applied the EMD algorithm to single trips independently. This means that if a driver performs a specific manoeuvre only once during the trip, the EMD will not recognise it as a motif since it only appears once. However, if we were to expand the search for motifs to a varied set of trips instead of focusing on a single trip, we expect to see these cases become very rare. Thus, this work should be further extended to multi-trip motif discovery. 

Finally, we used the lateral and longitudinal acceleration time-series separately to find simple manoeuvres such as turns and brakes. However, the ideal setup would be to look for motifs in the two acceleration axis at the same time and also include other sensors such as velocity. This way, we would be able to extract more complex manoeuvres and we would reduce errors created but changes in road inclination or blasts of wind.

This idea is equivalent to the task of detecting motifs in multidimensional time-series. The two main ways of tackling this problem are to reduce the number of variables to one and apply a motif discovery algorithm to the resulting 1-dimensional time-series \citep{tanaka_discovery_2005} or to apply a motif discovery algorithm to each individual dimension and search for co-occurrences of the 1-dimensional motifs to extract the final multidimensional motifs \citep{vahdatpour_toward_2009, minnen_detecting_2007, balasubramanian_discovering_2016}. While the first has the advantage of avoiding running a motif detection algorithm in all the dimensions, the second is more accurate as there is no information loss due to dimensionality reduction. Therefore, it would be very important to explore these two options and to discover if they are suited for the use-case of manoeuvre detection.

\bibliographystyle{unsrtnat} 
\bibliography{references}  

\begin{thebibliography}{36}
\providecommand{\natexlab}[1]{#1}
\providecommand{\url}[1]{\texttt{#1}}
\expandafter\ifx\csname urlstyle\endcsname\relax
  \providecommand{\doi}[1]{doi: #1}\else
  \providecommand{\doi}{doi: \begingroup \urlstyle{rm}\Url}\fi

\bibitem[Dingus et~al.(2016)Dingus, Guo, Lee, Antin, Perez, Buchanan-King, and
  Hankey]{dingus_driver_2016}
Thomas~A. Dingus, Feng Guo, Suzie Lee, Jonathan~F. Antin, Miguel Perez, Mindy
  Buchanan-King, and Jonathan Hankey.
\newblock Driver crash risk factors and prevalence evaluation using
  naturalistic driving data.
\newblock \emph{Proceedings of the National Academy of Sciences}, 113\penalty0
  (10):\penalty0 2636--2641, March 2016.
\newblock ISSN 0027-8424, 1091-6490.
\newblock \doi{10.1073/pnas.1513271113}.
\newblock URL \url{http://www.pnas.org/content/113/10/2636}.

\bibitem[Dozza(2013)]{dozza_what_2013}
Marco Dozza.
\newblock What factors influence drivers’ response time for evasive maneuvers
  in real traffic?
\newblock \emph{Accident Analysis \& Prevention}, 58:\penalty0 299--308,
  September 2013.
\newblock ISSN 0001-4575.
\newblock \doi{10.1016/j.aap.2012.06.003}.
\newblock URL
  \url{http://www.sciencedirect.com/science/article/pii/S0001457512002254}.

\bibitem[Carsten et~al.(2013)Carsten, Kircher, and
  Jamson]{carsten_vehicle-based_2013}
Oliver Carsten, Katja Kircher, and Samantha Jamson.
\newblock Vehicle-based studies of driving in the real world: {The} hard truth?
\newblock \emph{Accident Analysis \& Prevention}, 58:\penalty0 162--174,
  September 2013.
\newblock ISSN 0001-4575.
\newblock \doi{10.1016/j.aap.2013.06.006}.
\newblock URL
  \url{http://www.sciencedirect.com/science/article/pii/S0001457513002340}.

\bibitem[Tselentis et~al.(2017)Tselentis, Yannis, and
  Vlahogianni]{tselentis_innovative_2017}
Dimitrios~I. Tselentis, George Yannis, and Eleni~I. Vlahogianni.
\newblock Innovative motor insurance schemes: {A} review of current practices
  and emerging challenges.
\newblock \emph{Accident Analysis \& Prevention}, 98:\penalty0 139--148,
  January 2017.
\newblock ISSN 0001-4575.
\newblock \doi{10.1016/j.aap.2016.10.006}.
\newblock URL
  \url{http://www.sciencedirect.com/science/article/pii/S0001457516303670}.

\bibitem[Johnson and Trivedi(2011)]{johnson_driving_2011}
D.~A. Johnson and M.~M. Trivedi.
\newblock Driving style recognition using a smartphone as a sensor platform.
\newblock In \emph{2011 14th {International} {IEEE} {Conference} on
  {Intelligent} {Transportation} {Systems} ({ITSC})}, pages 1609--1615, October
  2011.
\newblock \doi{10.1109/ITSC.2011.6083078}.

\bibitem[Paefgen et~al.(2012)Paefgen, Kehr, Zhai, and
  Michahelles]{paefgen_driving_2012}
Johannes Paefgen, Flavius Kehr, Yudan Zhai, and Florian Michahelles.
\newblock Driving {Behavior} {Analysis} with {Smartphones}: {Insights} from a
  {Controlled} {Field} {Study}.
\newblock In \emph{Proceedings of the 11th {International} {Conference} on
  {Mobile} and {Ubiquitous} {Multimedia}}, {MUM} '12, pages 36:1--36:8, New
  York, NY, USA, 2012. ACM.
\newblock ISBN 978-1-4503-1815-0.
\newblock \doi{10.1145/2406367.2406412}.
\newblock URL \url{http://doi.acm.org/10.1145/2406367.2406412}.

\bibitem[Kantor and Stárek(2014)]{kantor_design_2014}
S.~Kantor and T.~Stárek.
\newblock Design of {Algorithms} for {Payment} {Telematics} {Systems}
  {Evaluating} {Driver}'s {Driving} {Style}.
\newblock \emph{Transactions on Transport Sciences}, 7\penalty0 (1):\penalty0
  9--16, March 2014.
\newblock ISSN 1802971X, 18029876.
\newblock \doi{10.2478/v10158-012-0049-5}.
\newblock URL \url{http://tots.upol.cz/doi/10.2478/v10158-012-0049-5.html}.

\bibitem[Eren et~al.(2012)Eren, Makinist, Akin, and
  Yilmaz]{eren_estimating_2012}
H.~Eren, S.~Makinist, E.~Akin, and A.~Yilmaz.
\newblock Estimating driving behavior by a smartphone.
\newblock In \emph{2012 {IEEE} {Intelligent} {Vehicles} {Symposium}}, pages
  234--239, June 2012.
\newblock \doi{10.1109/IVS.2012.6232298}.

\bibitem[Saleh et~al.(2017)Saleh, Hossny, and Nahavandi]{saleh_driving_2017}
K.~Saleh, M.~Hossny, and S.~Nahavandi.
\newblock Driving behavior classification based on sensor data fusion using
  {LSTM} recurrent neural networks.
\newblock In \emph{2017 {IEEE} 20th {International} {Conference} on
  {Intelligent} {Transportation} {Systems} ({ITSC})}, pages 1--6, October 2017.
\newblock \doi{10.1109/ITSC.2017.8317835}.

\bibitem[Weidner et~al.(2016)Weidner, Transchel, and
  Weidner]{weidner_classification_2016}
W.~Weidner, F.~W.~G. Transchel, and R.~Weidner.
\newblock Classification of scale-sensitive telematic observables for
  riskindividual pricing.
\newblock \emph{European Actuarial Journal}, 6\penalty0 (1):\penalty0 3--24,
  July 2016.
\newblock ISSN 2190-9733, 2190-9741.
\newblock \doi{10.1007/s13385-016-0127-x}.
\newblock URL
  \url{https://link.springer.com/article/10.1007/s13385-016-0127-x}.

\bibitem[Murphey et~al.(2009)Murphey, Milton, and
  Kiliaris]{murphey_drivers_2009}
Y.~L. Murphey, R.~Milton, and L.~Kiliaris.
\newblock Driver's style classification using jerk analysis.
\newblock In \emph{2009 {IEEE} {Workshop} on {Computational} {Intelligence} in
  {Vehicles} and {Vehicular} {Systems}}, pages 23--28, March 2009.
\newblock \doi{10.1109/CIVVS.2009.4938719}.

\bibitem[Xie et~al.(2018)Xie, Hilal, and Kulić]{xie_driving_2018}
J.~Xie, A.~R. Hilal, and D.~Kulić.
\newblock Driving {Maneuver} {Classification}: {A} {Comparison} of {Feature}
  {Extraction} {Methods}.
\newblock \emph{IEEE Sensors Journal}, 18\penalty0 (12):\penalty0 4777--4784,
  June 2018.
\newblock ISSN 1530-437X.
\newblock \doi{10.1109/JSEN.2017.2780089}.

\bibitem[Singh et~al.(2017)Singh, Bansal, and Sofat]{singh_smartphone_2017}
Gurdit Singh, Divya Bansal, and Sanjeev Sofat.
\newblock A smartphone based technique to monitor driving behavior using {DTW}
  and crowdsensing.
\newblock \emph{Pervasive and Mobile Computing}, 40:\penalty0 56--70, September
  2017.
\newblock ISSN 1574-1192.
\newblock \doi{10.1016/j.pmcj.2017.06.003}.
\newblock URL
  \url{http://www.sciencedirect.com/science/article/pii/S1574119216301250}.

\bibitem[Woo and Kulić(2016)]{woo_manoeuvre_2016}
C.~Woo and D.~Kulić.
\newblock Manoeuvre segmentation using smartphone sensors.
\newblock In \emph{2016 {IEEE} {Intelligent} {Vehicles} {Symposium} ({IV})},
  pages 572--577, June 2016.
\newblock \doi{10.1109/IVS.2016.7535444}.

\bibitem[Camlica et~al.(2016)Camlica, Hilal, and Kulić]{camlica_feature_2016}
Z.~Camlica, A.~Hilal, and D.~Kulić.
\newblock Feature abstraction for driver behaviour detection with stacked
  sparse auto-encoders.
\newblock In \emph{2016 {IEEE} {International} {Conference} on {Systems},
  {Man}, and {Cybernetics} ({SMC})}, pages 003299--003304, October 2016.
\newblock \doi{10.1109/SMC.2016.7844743}.

\bibitem[Wu et~al.(2016)Wu, Zhang, and Dong]{wu_novel_2016}
Minglin Wu, Sheng Zhang, and Yuhan Dong.
\newblock A {Novel} {Model}-{Based} {Driving} {Behavior} {Recognition} {System}
  {Using} {Motion} {Sensors}.
\newblock \emph{Sensors}, 16\penalty0 (10):\penalty0 1746, October 2016.
\newblock \doi{10.3390/s16101746}.
\newblock URL \url{https://www.mdpi.com/1424-8220/16/10/1746}.

\bibitem[Júnior et~al.(2017)Júnior, Carvalho, Ferreira, Souza, Suhara,
  Pentland, and Pessin]{junior_driver_2017}
Jair~Ferreira Júnior, Eduardo Carvalho, Bruno~V. Ferreira, Cleidson~de Souza,
  Yoshihiko Suhara, Alex Pentland, and Gustavo Pessin.
\newblock Driver behavior profiling: {An} investigation with different
  smartphone sensors and machine learning.
\newblock \emph{PLOS ONE}, 12\penalty0 (4):\penalty0 e0174959, 2017.
\newblock ISSN 1932-6203.
\newblock \doi{10.1371/journal.pone.0174959}.
\newblock URL
  \url{https://journals.plos.org/plosone/article?id=10.1371/journal.pone.0174959}.

\bibitem[Keogh and Lin(2005)]{keogh_clustering_2005}
Eamonn Keogh and Jessica Lin.
\newblock Clustering of time-series subsequences is meaningless: implications
  for previous and future research.
\newblock \emph{Knowledge and Information Systems}, 8\penalty0 (2):\penalty0
  154--177, August 2005.
\newblock ISSN 0219-3116.
\newblock \doi{10.1007/s10115-004-0172-7}.
\newblock URL \url{https://doi.org/10.1007/s10115-004-0172-7}.

\bibitem[Torkamani and Lohweg(2017)]{torkamani_survey_2017}
Sahar Torkamani and Volker Lohweg.
\newblock Survey on time series motif discovery.
\newblock \emph{Wiley Interdisciplinary Reviews: Data Mining and Knowledge
  Discovery}, 7\penalty0 (2):\penalty0 e1199, 2017.
\newblock ISSN 1942-4795.
\newblock \doi{10.1002/widm.1199}.
\newblock URL \url{https://onlinelibrary.wiley.com/doi/abs/10.1002/widm.1199}.

\bibitem[Mueen(2014)]{mueen_time_2014}
Abdullah Mueen.
\newblock Time series motif discovery: dimensions and applications.
\newblock \emph{Wiley Interdisciplinary Reviews: Data Mining and Knowledge
  Discovery}, 4\penalty0 (2):\penalty0 152--159, 2014.
\newblock ISSN 1942-4795.
\newblock \doi{10.1002/widm.1119}.
\newblock URL \url{https://onlinelibrary.wiley.com/doi/abs/10.1002/widm.1119}.

\bibitem[Tanaka et~al.(2005)Tanaka, Iwamoto, and Uehara]{tanaka_discovery_2005}
Yoshiki Tanaka, Kazuhisa Iwamoto, and Kuniaki Uehara.
\newblock Discovery of {Time}-{Series} {Motif} from {Multi}-{Dimensional}
  {Data} {Based} on {MDL} {Principle}.
\newblock \emph{Machine Learning}, 58\penalty0 (2):\penalty0 269--300, February
  2005.
\newblock ISSN 1573-0565.
\newblock \doi{10.1007/s10994-005-5829-2}.
\newblock URL \url{https://doi.org/10.1007/s10994-005-5829-2}.

\bibitem[Lin et~al.(2002)Lin, Keogh, Lonardi, and Patel]{lin_finding_2002}
Jessica Lin, Eamonn Keogh, Stefano Lonardi, and Pranav Patel.
\newblock Finding {Motifs} in {Time} {Series}.
\newblock In \emph{Proceedings of the {Second} {Workshop} on {Temporal} {Data}
  {Mining}}, pages 53--68, July 2002.
\newblock URL \url{http://citeseer.ist.psu.edu/lin02finding.html}.

\bibitem[Lin et~al.(2007)Lin, Keogh, Wei, and Lonardi]{lin_experiencing_2007}
Jessica Lin, Eamonn Keogh, Li~Wei, and Stefano Lonardi.
\newblock Experiencing {SAX}: a novel symbolic representation of time series.
\newblock \emph{Data Mining and Knowledge Discovery}, 15\penalty0 (2):\penalty0
  107--144, October 2007.
\newblock ISSN 1384-5810, 1573-756X.
\newblock \doi{10.1007/s10618-007-0064-z}.
\newblock URL
  \url{https://link.springer.com/article/10.1007/s10618-007-0064-z}.

\bibitem[Agrawal et~al.(1993)Agrawal, Faloutsos, and
  Swami]{agrawal_efficient_1993}
Rakesh Agrawal, Christos Faloutsos, and Arun Swami.
\newblock Efficient similarity search in sequence databases.
\newblock In \emph{Foundations of {Data} {Organization} and {Algorithms}},
  Lecture {Notes} in {Computer} {Science}, pages 69--84. Springer, Berlin,
  Heidelberg, October 1993.
\newblock ISBN 978-3-540-57301-2 978-3-540-48047-1.
\newblock \doi{10.1007/3-540-57301-1_5}.
\newblock URL \url{https://link.springer.com/chapter/10.1007/3-540-57301-1_5}.

\bibitem[Das et~al.(1998)Das, Lin, Mannila, Renganathan, and
  Smyth]{das_rule_1998}
Gautam Das, King-Ip Lin, Heikki Mannila, Gopal Renganathan, and Padhraic Smyth.
\newblock Rule {Discovery} from {Time} {Series}.
\newblock In \emph{Proceedings of the {Fourth} {International} {Conference} on
  {Knowledge} {Discovery} and {Data} {Mining}}, {KDD}'98, pages 16--22, New
  York, NY, 1998. AAAI Press.
\newblock URL \url{http://dl.acm.org/citation.cfm?id=3000292.3000296}.

\bibitem[Berndt and Clifford(1994)]{berndt_using_1994}
Donald~J. Berndt and James Clifford.
\newblock Using {Dynamic} {Time} {Warping} to {Find} {Patterns} in {Time}
  {Series}.
\newblock In \emph{Proceedings of the {AAAI} {Workshop} on {Knowledge}
  {Discovery} in {Databases}}, pages 359--370, 1994.

\bibitem[Nunthanid et~al.(2012)Nunthanid, Niennattrakul, and
  Ratanamahatana]{nunthanid_parameter-free_2012}
P.~Nunthanid, V.~Niennattrakul, and C.~A. Ratanamahatana.
\newblock Parameter-free motif discovery for time series data.
\newblock In \emph{2012 9th {International} {Conference} on {Electrical}
  {Engineering}/{Electronics}, {Computer}, {Telecommunications} and
  {Information} {Technology}}, pages 1--4, May 2012.
\newblock \doi{10.1109/ECTICon.2012.6254126}.

\bibitem[Gao and Lin(2018)]{gao_exploring_2018}
Yifeng Gao and Jessica Lin.
\newblock Exploring variable-length time series motifs in one hundred million
  length scale.
\newblock \emph{Data Mining and Knowledge Discovery}, 32\penalty0 (5):\penalty0
  1200--1228, September 2018.
\newblock ISSN 1573-756X.
\newblock \doi{10.1007/s10618-018-0570-1}.
\newblock URL \url{https://doi.org/10.1007/s10618-018-0570-1}.

\bibitem[Rissanen(1998)]{rissanen_stochastic_1998}
Jorma Rissanen.
\newblock \emph{Stochastic {Complexity} {In} {Statistical} {Inquiry}}.
\newblock World Scientific, October 1998.
\newblock ISBN 978-981-4507-40-0.
\newblock Google-Books-ID: KY4GCwAAQBAJ.

\bibitem[Ester et~al.(1996)Ester, Kriegel, Sander, and
  Xu]{ester_density-based_1996}
Martin Ester, Hans-Peter Kriegel, Jörg Sander, and Xiaowei Xu.
\newblock A {Density}-{Based} {Algorithm} for {Discovering} {Clusters} in
  {Large} {Spatial} {Databases} with {Noise}.
\newblock In \emph{Proceedings of the 2nd {International} {Conference} on
  {Knowledge} {Discovery} and {Data} {Mining}}, pages 226--231, 1996.

\bibitem[Romera et~al.(2016)Romera, Bergasa, and Arroyo]{romera_need_2016}
E.~Romera, L.~M. Bergasa, and R.~Arroyo.
\newblock Need data for driver behaviour analysis? {Presenting} the public
  {UAH}-{DriveSet}.
\newblock In \emph{2016 {IEEE} 19th {International} {Conference} on
  {Intelligent} {Transportation} {Systems} ({ITSC})}, pages 387--392, November
  2016.
\newblock \doi{10.1109/ITSC.2016.7795584}.

\bibitem[Bergasa et~al.(2014)Bergasa, Almería, Almazán, Yebes, and
  Arroyo]{bergasa_drivesafe:_2014}
L.~M. Bergasa, D.~Almería, J.~Almazán, J.~J. Yebes, and R.~Arroyo.
\newblock {DriveSafe}: {An} app for alerting inattentive drivers and scoring
  driving behaviors.
\newblock In \emph{2014 {IEEE} {Intelligent} {Vehicles} {Symposium}
  {Proceedings}}, pages 240--245, June 2014.
\newblock \doi{10.1109/IVS.2014.6856461}.

\bibitem[Romera et~al.(2015)Romera, Bergasa, and Arroyo]{romera_real-time_2015}
E.~Romera, L.~M. Bergasa, and R.~Arroyo.
\newblock A {Real}-{Time} {Multi}-scale {Vehicle} {Detection} and {Tracking}
  {Approach} for {Smartphones}.
\newblock In \emph{2015 {IEEE} 18th {International} {Conference} on
  {Intelligent} {Transportation} {Systems}}, pages 1298--1303, September 2015.
\newblock \doi{10.1109/ITSC.2015.213}.

\bibitem[Vahdatpour et~al.(2009)Vahdatpour, Amini, and
  Sarrafzadeh]{vahdatpour_toward_2009}
Alireza Vahdatpour, Navid Amini, and Majid Sarrafzadeh.
\newblock Toward {Unsupervised} {Activity} {Discovery} {Using}
  {Multi}-dimensional {Motif} {Detection} in {Time} {Series}.
\newblock In \emph{Proceedings of the 21st {International} {Jont} {Conference}
  on {Artifical} {Intelligence}}, {IJCAI}'09, pages 1261--1266, San Francisco,
  CA, USA, 2009. Morgan Kaufmann Publishers Inc.
\newblock URL \url{http://dl.acm.org/citation.cfm?id=1661445.1661647}.

\bibitem[Minnen et~al.(2007)Minnen, Isbell, Essa, and
  Starner]{minnen_detecting_2007}
D.~Minnen, C.~Isbell, I.~Essa, and T.~Starner.
\newblock Detecting {Subdimensional} {Motifs}: {An} {Efficient} {Algorithm} for
  {Generalized} {Multivariate} {Pattern} {Discovery}.
\newblock In \emph{Seventh {IEEE} {International} {Conference} on {Data}
  {Mining} ({ICDM} 2007)}, pages 601--606, October 2007.
\newblock \doi{10.1109/ICDM.2007.52}.

\bibitem[Balasubramanian et~al.(2016)Balasubramanian, Wang, and
  Prabhakaran]{balasubramanian_discovering_2016}
A.~Balasubramanian, J.~Wang, and B.~Prabhakaran.
\newblock Discovering {Multidimensional} {Motifs} in {Physiological} {Signals}
  for {Personalized} {Healthcare}.
\newblock \emph{IEEE Journal of Selected Topics in Signal Processing},
  10\penalty0 (5):\penalty0 832--841, August 2016.
\newblock ISSN 1932-4553.
\newblock \doi{10.1109/JSTSP.2016.2543679}.

\end{thebibliography}




\end{document}